\documentclass[sigconf]{acmart}
\AtBeginDocument{%
  }

\setcopyright{none}
\copyrightyear{2026}
\acmYear{2026}
\acmDOI{XXXXXXX.XXXXXXX}
\acmConference[Accepted to ACM SIGSPATIAL'26]{Accepted to the 34th ACM International Conference on Advances in Geographic Information Systems}{Nov}{2026}
\acmISBN{978-1-4503-XXXX-X/2018/06}
\setcopyright{none}
\renewcommand\footnotetextcopyrightpermission[1]{}

\usepackage{multirow}
\usepackage{colortbl}
\usepackage{arydshln}
\usepackage{xcolor}

\newcommand{\ub}[1]{\textcolor{gray}{#1}}

\begin{document}

\title[FAIRY: Smart-Agriculture Agentic Engine for Full-Season Soybean Farm Operations]{Deploying and Evaluating a Smart-Agriculture Agentic Engine for Full-Season Soybean Farm Operations}


\author{Ao Qu}
\authornote{Equal contribution.}
\email{aoqu@stu.hit.edu.cn}
\orcid{0009-0008-8230-3211}
\affiliation{%
  \department{Faculty of Computing} 
  \institution{Harbin Institute of Technology}
  \department{State Key Laboratory of Smart Farm Technologies and Systems}
  \city{Harbin}
  \country{China}
}

\author{Panagiotis Michelakis}
\authornotemark[1]
\authornote{Currently with new affiliation.}
\email{panosg@synkrasis-labs.com}
\orcid{0009-0003-0498-0499}
\affiliation{%
  \department{School of Electrical and Computer Engineering}
  \institution{National Technical University of Athens}
  \city{Athens}
  \state{}
  \country{Greece}
}

\author{Linyuan Han}
\authornotemark[1]
\email{linyuanhan26@stu.hit.edu.cn}
\orcid{0009-0009-2041-7202}
\affiliation{%
  \department{Faculty of Computing}
  \institution{Harbin Institute of Technology}
  \department{State Key Laboratory of Smart Farm Technologies and Systems}
  \city{Harbin}
  \country{China}
}

\author{Yiannis Hadjiyianni}
\email{yiannisha@synkrasis-labs.com}
\orcid{0009-0003-2413-6375}
\affiliation{%
  \department{School of Electrical and Computer Engineering}
  \institution{National Technical University of Athens}
  \city{Athens}
  \state{}
  \country{Greece}
}

\author{Kun Ouyang}
\email{kunouyang@stu.hit.edu.cn}
\orcid{0009-0006-4754-9625}
\affiliation{%
  \department{Faculty of Computing}
  \institution{Harbin Institute of Technology}
  \department{State Key Laboratory of Smart Farm Technologies and Systems}
  \city{Harbin}
  \country{China}
}

\author{Konstantinos Siskos}
\email{siskos@synkrasis-labs.com}
\orcid{0009-0000-0596-2522}
\affiliation{%
  \department{School of Electrical and Computer Engineering}
  \institution{National Technical University of Athens}
  \city{Athens}
  \state{}
  \country{Greece}
}

\author{Feng Li}
\authornote{Corresponding authors.}
\email{feng.li@hit.edu.cn}
\orcid{0009-0008-3995-0931}
\affiliation{%
  \department{Faculty of Computing}
  \institution{Harbin Institute of Technology}
  \department{State Key Laboratory of Smart Farm Technologies and Systems}
  \city{Harbin}
  \country{China}
}

\author{Ran Meng}
\authornotemark[3]
\email{mengran@hit.edu.cn}
\orcid{0000-0003-4756-9934}
\affiliation{%
  \department{Faculty of Computing}
  \institution{Harbin Institute of Technology}
  \department{State Key Laboratory of Smart Farm Technologies and Systems}
  \city{Harbin}
  \country{China}
}

\author{Jingchi Jiang}
\authornotemark[3]
\email{jiangjingchi@hit.edu.cn}
\orcid{0000-0003-2167-4082}
\affiliation{%
  \department{Faculty of Computing}
  \institution{Harbin Institute of Technology}
  \department{State Key Laboratory of Smart Farm Technologies and Systems}
  \city{Harbin}
  \country{China}
}

\author{Dimitrios Stamoulis}
\authornote{Project Lead: \textbf{FAIRY Platform}, smart-farm multi-agent system and world models.}
\authornotemark[3]
\authornotemark[1]
\email{dimi@hit.edu.cn}
\orcid{0000-0003-1682-9350}
\affiliation{%
  \department{Faculty of Computing}
  \institution{Harbin Institute of Technology}
  \department{State Key Laboratory of Smart Farm Technologies and Systems}
  \city{Harbin}
  \country{China}
}

\author{Jie Liu}
\authornotemark[3]
\email{jieliu@hit.edu.cn}
\orcid{0000-0001-6209-6886}
\affiliation{%
  \department{Faculty of Computing}
  \institution{Harbin Institute of Technology}
  \department{State Key Laboratory of Smart Farm Technologies and Systems}
  \city{Harbin}
  \country{China}
}

\renewcommand{\shortauthors}{Qu et al.}

\begin{abstract}
This paper presents FAIRY, a full-stack smart-agriculture agent system developed for and deployed to an operating soybean research farm at Harbin Institute of Technology's smart-agriculture site. We develop FAIRY to execute and evaluate agentic agronomic operations on full-season \emph{spatiotemporal} workflows that span ridge preparation, planting, irrigation, fertilization, pest and disease treatment, harvest, grain handling, drying, and storage. FAIRY integrates APIs and infrastructure across production-grade machinery, fixed soil and canopy sensors, multispectral and thermal drones, satellite vegetation products, a weather station, calibrated crop-process models, agronomic records, and multi-season yield histories. The system is built around the novel ``everything is an event'' execution paradigm, which represents spatiotemporal world evolution, remote sensing and UAV observations, sensor readings, crop-growth transitions, machinery actions, and management interventions as state-changing events in a shared farm process engine. On top of this event-driven \textit{world model}, FAIRY implements a complete agentic stack: a knowledge library of atomic agronomic skills; multi-agent controller and orchestration backends; frontier- and edge-model execution; full-path trace logging; and deployment profiling on local nodes. We use FAIRY to evaluate nine state-of-the-art agent controllers across one hundred full-season soybean scenarios that preserve the operational coupling between spatial observations in a 64-ridge field, temporal decision sequences, agronomic constraints, delayed effects, and final yield. We develop an evaluation suite that combines agentic success, \textit{full-path spatiotemporal} correctness, token cost, and edge-device runtime. Our results show that our \textit{spatiotemporally} grounded Kendall correctness (KTC) improves alignment with downstream yield outcomes compared to existing order-only and exact-match metrics. Our analyses further show that hierarchical agronomic skills and expert operational context substantially improve long-horizon behavior compared to geospatial in-context learning and LLM-as-an-Expert schemes.
\end{abstract}



\maketitle

\section{Introduction}

Recent advances in agentic AI have made tool-augmented language models increasingly useful for geospatial analysis, where tasks often require data discovery, API selection, model invocation, map operations, and multi-step interpretation over spatial and temporal data. Prior geospatial agent systems have shown this potential across remote sensing, Earth observation, urban analysis, forestry, climate studies, agriculture monitoring, and satellite-vision workflows~\cite{lee2025multiagent,Bhattaram2025geoflow}. As reflected by recent releases in commercial geospatial platforms, such as Google Earth AI and Microsoft Planetary Computer offerings for agriculture and environmental monitoring, these settings are a natural fit for agents because geospatial workflows require selecting appropriate imagery or products, applying spatial filters, invoking detection or classification models, reasoning over intermediate outputs, and producing map-based results.

\begin{figure*}[ht!]
  \centering
  \includegraphics[width=\linewidth]{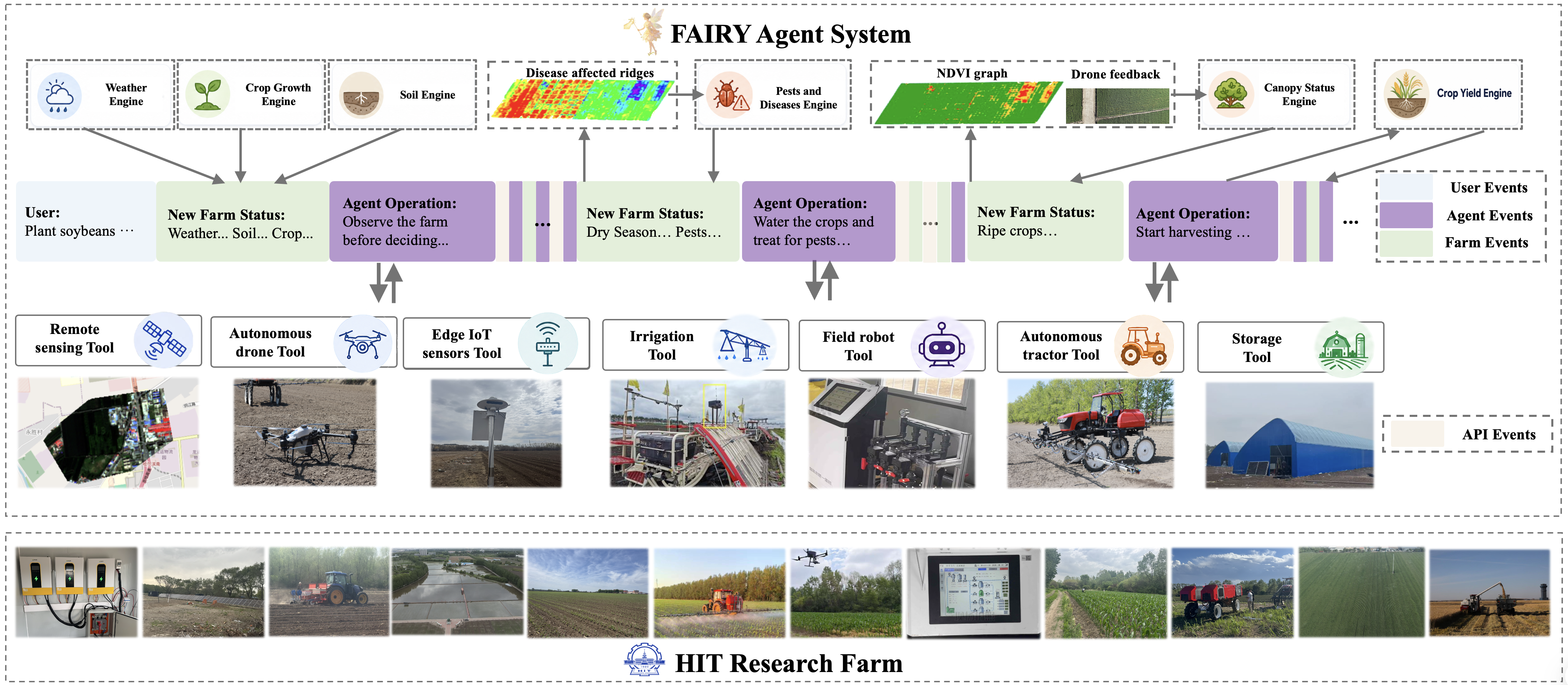}
  \caption{FAIRY deploys and evaluates
    agentic controllers over a full soybean season (planting through grain
    storage) on a 64-ridge operating research farm, coupling spatial
    observations, temporal decisions, delayed agronomic effects, and final
    yield.}
  \label{fig:FAIRY_overview}
  \Description{Full-process overview of the FAIRY agentic farm system.}
\end{figure*}

Despite these advances, deploying agentic systems for agricultural operations and complex agronomic spatiotemporal workflows remains difficult. Farm operations differ in three concrete ways~\cite{yan2026agrieval, zhang2026agriworldaworldtoolsprotocol,
xu2025multimodalagriculturalagentarchitecture, Seo_2026,
arregui2025allllmbasedheterogeneousmission,qu2026fullseason}: \textbf{Feedback is delayed:} a poor planting decision surfaces as weak emergence days later; a missed irrigation window appears as yield loss months later. \textbf{Observations are partial and spatial:} sensors are zone-level across ridges, drones are weather-gated, and ground inspection is sparse, so the agent reasons over a partially observed \emph{spatial} field. \textbf{Consequences propagate:} operational errors compound across the season, so an action can be syntactically valid and operationally wrong, such as harvesting before grain moisture is suitable.

At the same time, practical deployments of geospatial agents often inherit orchestration and evaluation practices originally developed for domains such as web automation, coding, or OS control, where feedback is immediate and the action space is bounded by application APIs. However, unlike cloud-centric remote-sensing and geospatial big-data pipelines, agricultural applications introduce a broader evaluation surface: Crop state evolves through biological growth, weather, soil-water dynamics, sensing constraints, and management interventions, so agent decisions are coupled to physical processes that unfold over days, weeks, and seasons. Ultimately, agent performance depends not only on successful tool calls, but also on spatial coverage, temporal alignment, data-product choice, and the operational meaning of the produced analysis.

In this work, we present a full-stack smart-agriculture agent system developed for an operating soybean research farm at a university smart-agriculture research site (Figure~\ref{fig:FAIRY_overview}) spanning a 64-ridge field. Real-life field operations span ridge preparation, planting, irrigation, fertilization, pest and disease treatment, harvest, grain handling, drying, and storage. The site has production-grade machinery; a digitized sensing layer covering fixed soil and canopy sensors, multispectral and thermal drones, satellite vegetation products, and a weather station; a calibrated, physics-grounded process model; and multi-season historical harvest and yield records. We are preparing a controlled portion of this field to be operated by agents alongside the existing human-operated workflow, with the goal of comparing agent-managed and human-managed operations in upcoming harvest seasons. This makes the evaluation question \textbf{practical}: we need to understand how agent systems behave on real farm operations before deciding what to deploy in the field.

To this end, we develop FAIRY as a full-stack agentic research engine for the operating farm. FAIRY first integrates the farm-facing infrastructure needed for deployment: production-grade machinery, fixed soil and canopy sensors, multispectral and thermal drones, satellite vegetation products, a weather station, calibrated crop-process models, agronomic records, and multi-season yield histories. The system then organizes these components through an ``everything is an event'' execution model, where weather updates, remote-sensing observations, UAV inspections, sensor readings, crop-growth transitions, machinery actions, and management interventions are represented as state-changing events in a shared farm process engine. On top of this event-driven \textit{world model}, FAIRY implements the agentic stack used in our evaluation: a retrieval-based knowledge library of atomic agronomic skills, nine agent controller backends with optional agent-to-agent (A2A) orchestration, frontier- and edge-model execution, full-path trace logging, and edge-device deployment profiling.

Before deploying agents on the controlled farm portion, we need an evaluation setting that can exercise farm operations beyond the limited number of observed historical seasons. The calibrated farm world model allows us to construct realistic scenarios that remain tied to our field geometry, crop-process assumptions, sensing layer, and operational workflow, while also covering conditions that have not yet occurred in the deployment record. Specifically, we build one hundred scenarios across three levels of complexity: atomic tasks, episode chains, and full-season scenarios. We develop a comprehensive evaluation suite which covers task success, temporal correctness, coverage, full-path correctness, and yield preservation. We conduct extensive metric-calibration studies that assess which trace-level metric best tracks yield, identifying temporally grounded Kendall correctness (KTC) as the best-calibrated predictor and exposing the failure modes of order-only and exact-match alternatives. Finally, we include an operational demonstration that runs the same event-driven runtime end-to-end, from a user request through drone and satellite observation to a geospatial visualization in the FAIRY interface.

Our results lead to four practical lessons for deploying agents in farm operations. First, domain expertise is the dominant lever: expert operational context reduces the full-season yield shortfall from $\sim$22\% under zero context to $\sim$5\% for Qwen on held-out L3 scenarios. Second, agent and farm performance diverge under longer horizons: short tasks are nearly solved, with $\geq$99\% temporal correctness and near-zero yield loss, while full-season scenarios remain materially below the human oracle. Third, multi-agent A2A orchestration introduces coordination cost that degrades both correctness and yield in our setting. Fourth, agriculture-tuned LLM context is only a partial substitute for human-expert context. We report these as field observations from an operating farm; the engine, scenarios, and evaluation are documented for reproduction.

\section{The FAIRY System: Architecture Overview}
\label{sec:FAIRY}

An overview of FAIRY is shown in Figure~\ref{fig:FAIRY_overview}. The system
couples an event-driven engine, a physics-grounded process stack, a tool/sensing layer, a knowledge library, and the agentic backend.

\subsection{Farm site and APIs}

\textbf{Farm equipment and sensing infrastructure.} The study site is an industry-grade university soybean research farm with a modeled $268\,\mathrm{m} \times 71\,\mathrm{m}$ field organized into 64 ridges, which serve as the atomic \emph{spatial} units for observation and intervention. We expose farm capabilities to the agent as function-calling tools derived from the installed APIs and operational interfaces. As summarized in Table~\ref{tab:farm_assets}, the site combines fixed soil, canopy, weather, light, radiation, and chlorophyll sensing; UAV and calibration assets for multispectral, thermal, and LiDAR observation; ground robotic platforms; production machinery; spraying equipment; and ridge-level irrigation and fertilization facilities. 

\begin{table}[h!]
\centering
\caption{FAIRY agent APIs closely model and connect to the underlying farm equipment and on-field sensing assets.}
\label{tab:farm_assets}
\small
\begin{tabular}{lll}
\toprule
\textbf{Category} & \textbf{Asset} & \textbf{Qty.} \\
\midrule
Field & Soybean ridges & 64 \\
Soil sensing & DF-G3012 + DF-HRS & 6 \\
Canopy sensing & Apogee canopy index sensor & 6 \\
Weather & WX-CQ10 weather station & 1 \\
Light sensing & DLS light sensor & 1 \\
Radiation & Solar-radiation sensor & 1 \\
Crop sensing & SPAD chlorophyll meter & 1 \\
UAV & DJI Matrice 4T thermal UAV & 1 \\
UAV & DJI Mavic 3 Multispectral & 1 \\
LiDAR & DJI Zenmuse L2 & 1 \\
UAV automation & DJI drone dock station & 2 \\
Calibration & Ground control markers & 12 \\
Calibration & P4M reflectance panels & 2 \\
Calibration & Blackbody radiation source & 1 \\
UAV sensing & Drone-mounted radiation sensor & 1 \\
Machinery & Tractor & 1 \\
Machinery & Container trailer & 1 \\
Spraying & Tractor-mounted spray boom & 1 \\
Spraying & Backpack spray tank & 1 \\
Irrigation/fertilization & Ridge-level facilities & per ridge \\
\bottomrule
\end{tabular}
\end{table}

\textbf{Satellite imagery and map products.} We use Google Earth Engine (GEE) to retrieve Sentinel-2 Surface Reflectance Harmonized imagery over the target area of interest (AOI), which we resolve either from a GEE asset or from a local AOI ZIP/shapefile. The retrieval pipeline filters scenes by AOI, date range, and cloudy-pixel percentage, sorts candidate scenes by cloud coverage, and exports the selected image as a 10 m GeoTIFF in EPSG:4326. We export bands B2, B3, B4, B5, B6, B7, B8, B8A, B11, and B12 through GEE export and Google Drive/API download logic. Downstream, we clip the raster to the AOI and georegister map placement from the GeoTIFF bounds in EPSG:4326. For visualization, we generate a satellite backdrop PNG from B4/B3/B2 RGB bands using percentile stretching and transparency masking, then overlay the result in the Leaflet-based map UI.

\textbf{Satellite crop classification.} We implement crop classification with an in-house XGBoost multiclass model. The classifier uses the \texttt{gbtree} booster with a \texttt{multi:softmax} objective and takes the ten Sentinel-2 bands together with vegetation indices including NDVI, kNDVI, GW1, GW2, LSWI, NDWI, EVI, EVI2, MSAVI, GNDVI, NDRE, GWCCI, and REP. We extract labeled pixels from crop polygons using geometry masks and use a stratified 70/30 train-test split. The model supports both training from scratch and loading a pretrained checkpoint with optional fine-tuning; in this study, we do not enable additional Heilongjiang regional fine-tuning. The operational output classes are rice, maize, soybean, other, and background. 

\begin{figure}[h!]
  \centering
  \includegraphics[width=\linewidth]{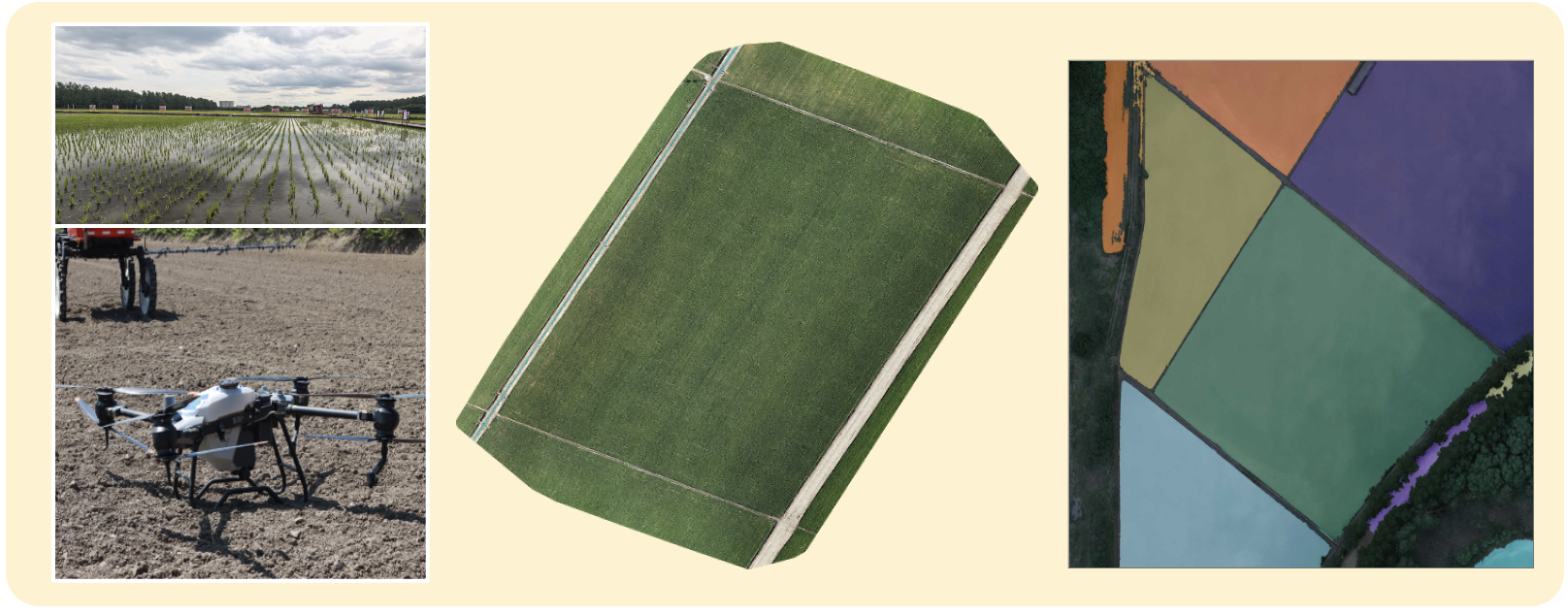}
  \caption{On-field drone inspections and UAV APIs.}
  \Description{Example visualizations of on-field drone inspections.}
\label{fig:drone_observation}
\end{figure}

\textbf{Drone imagery and APIs.} We collect field imagery at regular intervals with farm personnel and are also experimenting with direct UAV control through DJI Cloud API. The current deployment uses a DJI M300-RTK equipped with a DJI P1 full-frame RGB camera, producing 8192$\times$5460 imagery for field inspection. We generate orthomosaic products through DJI Terra, with OpenDroneMap also supported as an agent-callable processing backend. For field identification, we extract field-edge information using a multi-self-adaptive-threshold Canny operator and use the excess green index (ExG) to filter plant-covered regions. For missing-seedling detection, we combine overall plant density with local plant density estimated through sliding-window counting to identify sparse or missing-seedling areas. The drone and examples of the imagery and collected results are illustrated in Figure~\ref{fig:drone_observation}.

\subsection{Growth process dynamics}

\textbf{Weather model.} The farm world model is driven by weather and remote-sensing observations that define the external conditions for crop growth and management. Weather is generated through a WGEN/Richardson-style daily weather model~\cite{siler2022optimal}, providing precipitation, temperature, radiation, wind, and humidity at the resolution required by the farm event loop. The observation layer incorporates the remote-sensing and drone-based products available during operation. We treat these observations as state updates that provide spatial evidence about canopy vigor, water stress, and anomaly regions for follow-up inspection, irrigation, fertilization, or pest-management decisions.

\textbf{Growth model.} Plant growth dynamics are represented through a coupled soil, phenology, canopy, and biotic-pressure stack following~\cite{wang2025agrikwoo}. The soil component maintains water and temperature states using a bucket-style water balance over precipitation, irrigation, runoff, drainage, and evapotranspiration. Phenology follows a GDD-based soybean development model with seed-type-specific maturity targets~\cite{akyuz2017developing}, supporting cultivar-specific assumptions and stage-dependent management rules. The canopy and biomass model follows Monteith-style radiation-use and light-interception principles~\cite{monteith1977climate,monsi2005factor}. The biotic-pressure component tracks weed, insect, and disease pressure as weather- and stage-dependent processes with treatment effects~\cite{steduto2009aquacrop}. 

\textbf{Action effects and yield.} Together, these components define a process-level state space in which \textit{action effects} depend on crop stage, soil condition, recent weather, and observed canopy state. Management actions modify the crop trajectory through delayed and stage-dependent effects: Planting establishes stand fraction and emergence timing; irrigation changes soil-water availability over subsequent days; fertilization affects nutrient stress and canopy development. At maturity, we employ a yield-recovery model to convert accumulated crop state into harvested grain~\cite{humburg2019soybeanHarvestLosses}. 

\begin{figure}[h!]
  \centering
  \includegraphics[width=\linewidth]{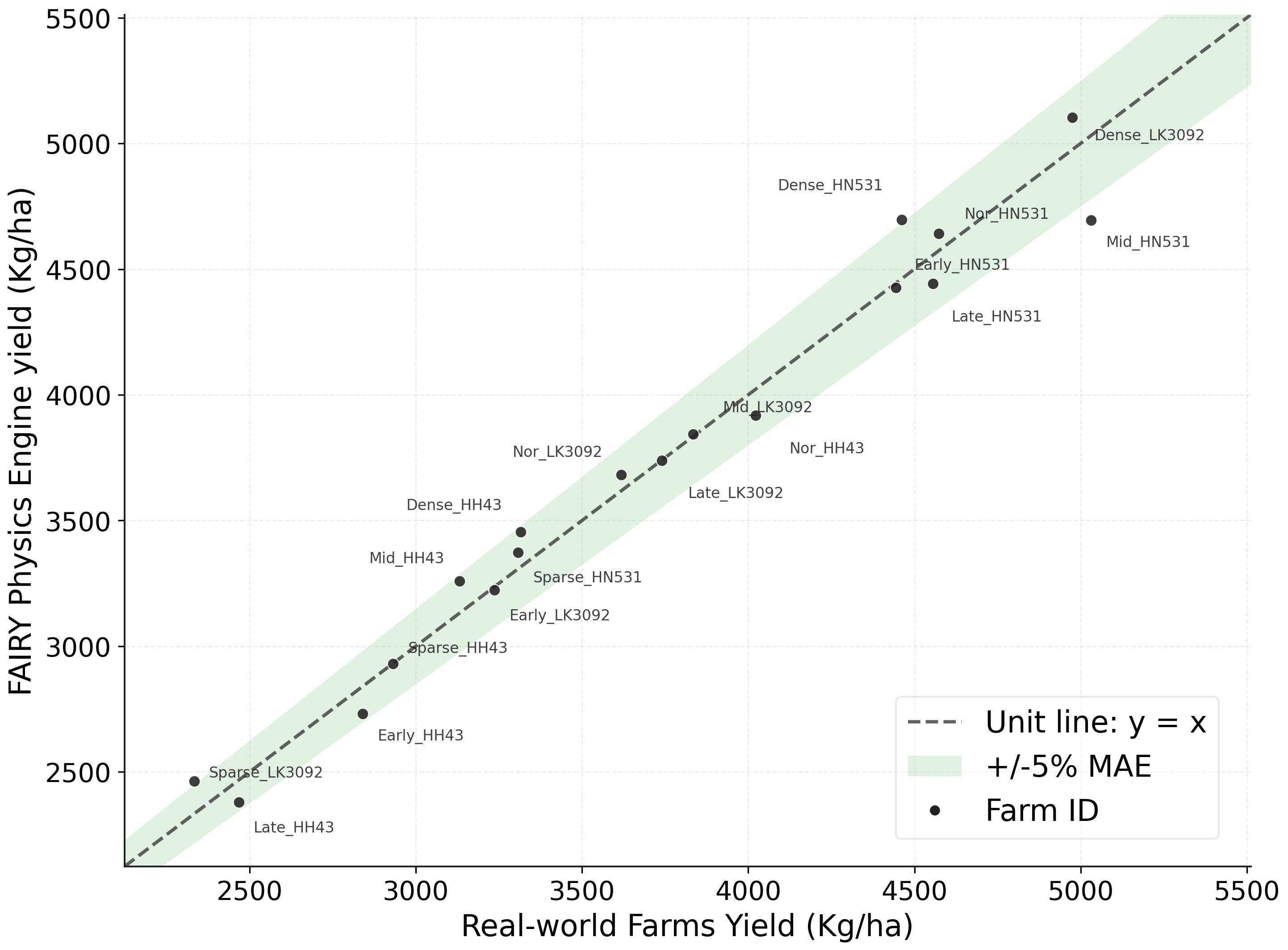}
  \caption{FAIRY Physics Engine yield vs. real-world farm yield across 18 Farm IDs.}
  \Description{FAIRY Physics Engine closely matches real-world farm yields from historical harvest data.}
\label{fig:growth_engine_validation}
\end{figure}

\textbf{Physics-engine calibration.} We validate the engine against historical data by reconstructing 18 plot-level soybean scenarios from the 2025 growing season (05-10 to 05-27) and harvest campaign (09-13 to 09-23). Different scenario IDs span different planting dates, density treatments, and cultivar type. Each FAIRY world model is initialized to closely match historical plot-specific management operations, cultivar type, planting density, fertilization schedule, and measured daily weather. We use observed phenology and yield records as evaluation targets, while the engine independently simulates emergence, crop growth, soil water and nutrient stress, biomass accumulation, harvest recovery, and final yield. As shown in Figure~\ref{fig:growth_engine_validation}, our engine yields closely track the observed yields across cultivars and treatments, with nearly all scenarios within yield prediction accuracies of $2\%$ MAE. We leave integration with commercial tools, such as WOFOST, to future work.

\subsection{Stateful event-driven engine}
We build the farm execution engine on the ARE framework~\cite{froger2025are},
which represents agent environments through stateful apps, an event queue,
notifications, scenarios, and logged execution traces. Farm operation is
naturally event-based: tools execute ridge preparation, planting, irrigation,
fertigation, pesticide application, harvest, and storage as timestamped farm
events, each with arguments (e.g., operation duration, target ridges) and
preconditions (e.g., weather and equipment readiness). The environment
maintains the farm state, event queue, notifications, and operation history,
while the controller observes and acts through the tool interface. Tool
invocation is emulated inside the digital twin, so operations retain their
farm-state effects and timing dependencies without real-clock execution.
This evaluation should therefore be interpreted as a deployment-readiness study rather than a completed agent-managed harvest trial. The physical infrastructure, sensing stack, machinery interfaces, and historical operation records are real, while the agent actions are replayed through the calibrated digital twin to evaluate safety and operational correctness before field deployment.

\subsection{Agent families and LLMs}
We integrate nine controller architectures and evaluate them over the scenario
suite: ReAct~\cite{yao2023react}, Plan-and-Act~\cite{erdogan2025planandact},
Reflexion~\cite{shinn2023reflexion}, AutoGen~\cite{wu2024autogen},
MMRL~\cite{tan2026multimodalMMRL}, ReWOO~\cite{xu2023rewoo},
LATS~\cite{zhou2024LATS}, CRITIC~\cite{gou2024critic}, and
GoT~\cite{Besta2024GoT}. Each controller supports two modes: \emph{direct}
tool access and \emph{agent-to-agent (A2A)}~\cite{a2aproject2025a2a}, where
weather, sensing, machinery, and operations queries are routed through
specialist app agents. Underlying farm state and tool capabilities are kept
consistent across controllers and modes. We run frontier API backends
(Qwen3.6-35B-A3B~\cite{yang2025qwen3technicalreport},
DeepSeek-V4-Flash~\cite{deepseekai2026deepseekv4}).
To reflect realistic deployment constraints, we also support edge execution through a vLLM API running on the NVIDIA Thor SDK with recent Qwen-3 and Gemma-4 model variants.

\subsection{Knowledge library}
\label{sec:library}
Farm management relies on tacit timing rules and operating discipline that are
usually implicit in human expertise. FAIRY encodes this knowledge as a
\textbf{knowledge library} of agronomic skills that controllers can
retrieve at decision time, mirroring hierarchical skill/prompt libraries used
for geospatial and remote-sensing agents~\cite{Bhattaram2025geoflow,
badmus2025powerchain__v2, singh2024geollm}. Each skill is a structured record with
an identifier, a natural-language title and description, a set of keywords, and
a reusable operation \emph{workflow} template. At decision time, a retriever
scores library entries against the current task and injects the top-$k$
($k{=}3$ by default) into the controller context. We expose two retrieval
mechanisms: a lightweight \emph{lexical} retriever scoring token overlap
between the query and each entry's searchable text, and a \emph{semantic}
retriever over sentence embeddings of the entire full-path workflow; both return ranked skills with scores.

This library underlies one of four interchangeable in-context regimes the
controller can run under: \textbf{Zero Context} (operation primitives only),
\textbf{LLM-as-an-Expert} (an agriculture-tuned LLM~\cite{wang2025agrikwoo}
generates scenario context from the task, tools, and world state),
\textbf{Skills Library} (retrieved skills injected into context), and
\textbf{Expert Instruct} (human-written scenario instructions). We vary how the library is \emph{organized} for retrieval (i.e., a flat pool versus a tier-grouped organization that separates atomic from composite operations)  and the \emph{ranking mechanism} used to select entries (i.e., manual selection, lexical text similarity, and path similarity).

\begin{table*}[ht!]
  \caption{Long-horizon analysis on atomic tasks (L1), episode
  chains (L2), and full-season scenarios (L3-test) for Qwen3.6-35B-A3B and DeepSeek-V4-Flash. Yield Loss is the \% drop from the human-oracle biological yield; Succ.\ is BFCL tool-call success; KTC is temporally grounded correctness; Token cost is reported as the average tokens per task and per agent API call.}
  \label{tab:horizon-api-results}
  \begin{center}
  \begin{tabular}{ll|cccccc|cccccc}
    \toprule
    Task & In-context & Yield & Succ. & Path & KTC & Tkns/ & Tkns/ & Yield & Succ. & Path & KTC & Tkns/ & Tkns/ \\
    Level & Learning   & Loss  & Score & Corr.&  Score & Task  & Step & Loss  & Score & Corr.& Score  & Task  & Step \\
    \midrule
    \rowcolor{black!10}
    & & \multicolumn{6}{c|}{\textit{Qwen3.6-35B-A3B}} & \multicolumn{6}{c}{\textit{DeepSeek-V4-Flash}} \\
    L1 & Zero Context    & 1.3\% & 42.0\% & 33.2\% & 75.1\% & 0.26M & 13.4k & 2.9\% & 46.1\% & 25.2\% & 78.1\% & 1.78M & 48.8k \\
       & LLM-as-an-Expert& 0.1\% & 86.6\% & 65.6\% & 98.7\% & 0.29M & 13.5k & 0.1\% & 89.0\% & 68.1\% & 98.8\% & 0.31M & 13.8k\\
       & Expert Instruct & 0.1\% & 85.6\% & 66.7\% & 99.2\% & 0.28M & 13.6k & 0.3\% & 88.2\% & 66.6\% & 99.2\% & 0.28M & 13.2k\\
    \midrule
    L2 & Zero Context    & 4.3\% & 39.6\% & 22.3\% & 73.9\% & 0.66M & 21.1k & 6.5\% & 48.6\% & 17.1\% & 74.3\% & 2.03M & 28.8k  \\
       & LLM-as-an-Expert& 0.8\% & 69.0\% & 49.4\% & 98.1\% & 0.92M & 27.1k & 0.9\% & 67.3\% & 45.3\% & 97.7\% & 2.28M & 45.0k \\
       & Expert Instruct & 0.8\% & 65.9\% & 51.2\% & 98.0\% & 1.20M & 32.7k & 0.8\% & 70.0\% & 52.4\% & 98.1\% & 1.41M & 33.8k \\
    \midrule
    L3 & Zero Context    & 22.3\% & 40.2\% & 22.8\% & 87.9\% & 3.89M & 33.0k & 12.6\% & 45.1\% & 28.5\% & 88.0\% & 5.55M & 40.5k  \\
       & LLM-as-an-Expert& 15.3\% & 46.3\% & 24.9\% & 89.8\% & 3.66M & 33.0k & 11.6\% & 52.9\% & 26.5\% & 89.0\% & 5.34M & 39.6k \\
       & Skills Library & 4.9\%  & 59.7\% & 30.9\% & 93.7\% & 4.09M & 37.4k & 6.3\%  & 65.7\% & 33.0\% & 93.6\% & 6.76M & 50.3k \\
       & Expert Instruct & 4.6\%  & 62.0\% & 28.6\% & 92.1\% & 2.93M & 29.0k & 2.9\%  & 67.6\% & 30.5\% & 92.7\% & 4.29M & 35.5k \\
    \bottomrule
  \end{tabular}
  \end{center}
\end{table*}

\subsection{Realistic full-season scenarios}
We curate scenarios from representative on-field procedures based on historical farm operations. We consider three levels of increasing complexity: \textbf{L1} atomic tasks require one or two tool calls (e.g., checking weather before a drone flight); \textbf{L2} episodes chain observation, diagnosis, and intervention (e.g., detecting an anomaly from a drone survey and applying targeted treatment); and \textbf{L3} full-season scenarios span planting through post-harvest storage, with multiple interventions, delayed consequences, and accumulated effects on crop state and recovered yield. For each scenario we follow the ARE annotation protocol~\cite{froger2025are}: three domain experts independently specify oracle solutions using the same operation primitives exposed to the agents; if the first two diverge we inspect the scenario to resolve ambiguity and revise, and the third expert confirms consistency. 

\vspace{+5pt}
\noindent
\fbox{%
\parbox{0.95\linewidth}{%
\textbf{Example scenarios by level:}

\textit{L1 (atomic).} ``The drone detected signs of aphids on ridges
15--25; verify the issue and apply an appropriate pesticide treatment.''
~\\
\textit{L2 (episode).} ``Remote sensing shows a localized low-NDVI area during
V4; diagnose drought vs.\ pest vs.\ nutrient deficiency and, if nutritional,
treat via ridge-level fertigation.'' 
~\\
\textit{L3 (full season).} ``Manage the full season from planting onward with weekly monitoring, addressing nutrient, water, pest, and disease issues from sensor/drone evidence, selecting a harvest window after maturity, and completing drying and storage.''
}%
}
\vspace{+5pt}

Replaying the oracle workflows in the engine produces human-oracle farm-state trajectories and target crop yields against which agent runs are compared. The evaluation reported here uses a full-season test set of 70 L3 scenarios, alongside the L1/L2 splits. We also construct a focused 20 L3-mini test set used for ablations, as well as a held-out validation set of 20 L3 scenarios for confirming the generalization of different library schemes. The oracle workflows are not assumed to be globally optimal; they represent expert-validated operational references for reproducible comparison. In future work, we will quantify inter-expert disagreement and test the sensitivity of KTC and Yield Loss rankings to alternative oracle choices.

\subsection{Evaluation suite and metrics}
\label{sec:metrics}
We evaluate each run along two axes: \emph{how the agent acted} (trace-level
correctness) and \emph{what it achieved} (yield outcome). Both compare an agent
run against the human-oracle workflow for the same scenario, represented as an
ordered set of timestamped farm events, each carrying a tool, its arguments,
target ridges, and a season-day.

\textbf{Trace-level correctness.} As our primary correctness metric we use
\textbf{KTC} (Kendall-style Temporal Correctness), a temporally grounded
full-path score that extends the path-correctness view
of~\cite{michelakis2025core} to farm-event order. After matching the agent's
executed operations to the oracle's, KTC measures the order agreement of the
matched operations as a normalized Kendall rank correlation,
$\mathrm{KTC}=(\tau+1)/2\in[0,1]$, rewarding required operations executed in
causally valid order and penalizing reordering. We complement KTC with two
reference trace metrics computed from the same (agent, oracle) pair:
\textbf{BFCL}, a tool-call success rate measured as set overlap of executed
$(\text{tool},\text{args})$ signatures (order- and time-agnostic);
\textbf{Path Correctness} (CORE), a normalized edit distance between the agent
and oracle operation sequences~\cite{michelakis2025core}.

\textbf{Agronomic outcome.} We report \textbf{Yield Loss}, the percentage shortfall of
the agent's biological yield relative to the human oracle
($1-\text{agent}_\text{bio}/\text{oracle}_\text{bio}$).
We pair correctness with outcome because they catch different failure modes:
trace metrics flag plausible-but-wrong sequences a final-state check would
miss~\cite{froger2025are}, while yield loss captures operational misjudgments
(e.g., harvesting on day 87 instead of 89) that a trace-only metric may treat
as minor deviations. We additionally report token cost and runtime per scenario.

\begin{figure}[ht!]
  \centering
  \includegraphics[width=\linewidth]{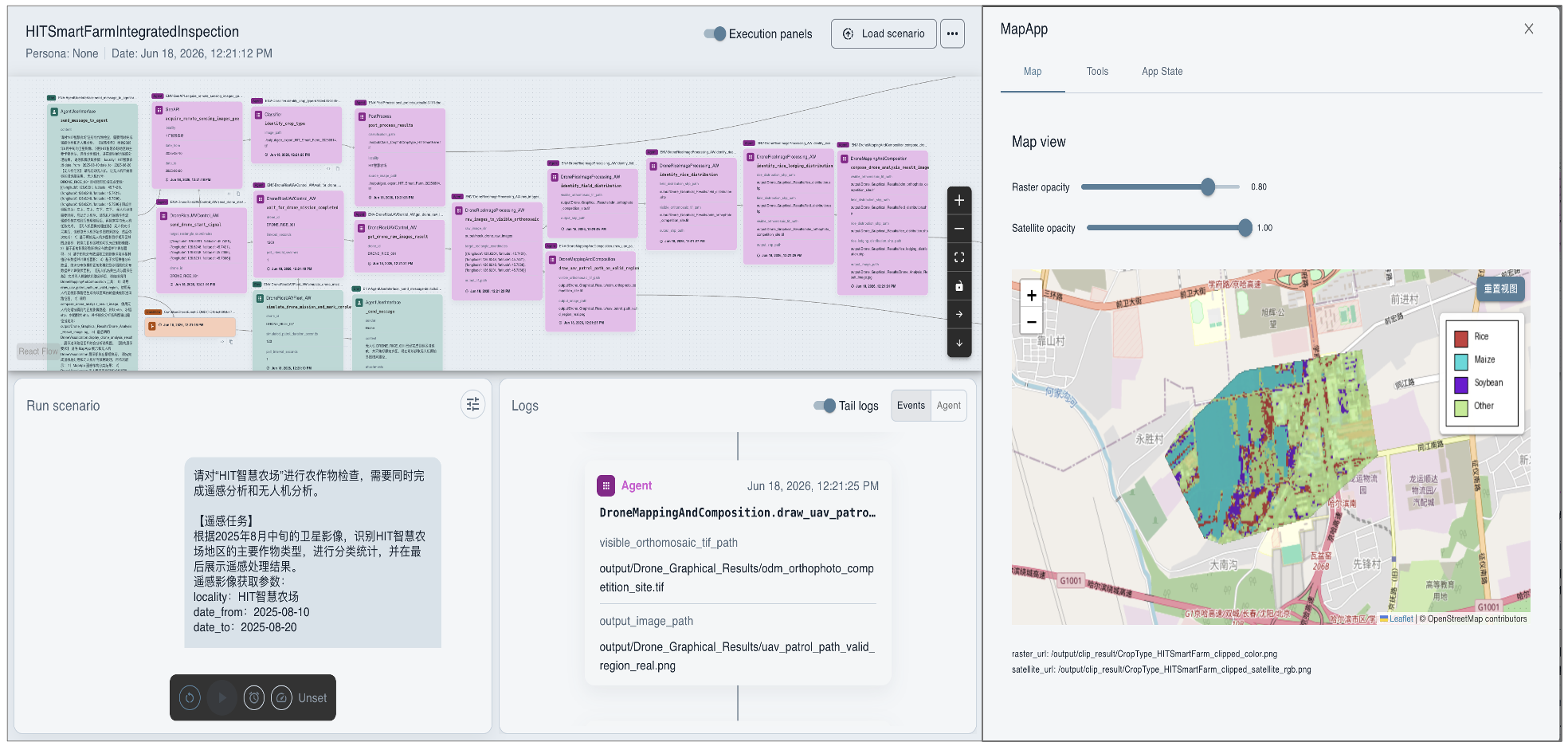}
\caption{FAIRY geospatial user interface. The web-based viewer displays the full-path agent workflow together with map-based visualizations.}
  \label{fig:FAIRY_ui}
  \Description{Operational demonstration of the agentic farm.}
\end{figure}

\subsection{FAIRY system user interface} 

We adapt the ARE user interface~\cite{froger2025are} from a generic agent-workflow environment into a geospatial inspection interface for agricultural workflows, as shown in Figure~\ref{fig:FAIRY_ui}. Notably, we extend the UI with domain-specific visualization support for satellite and UAV outputs, including zoomable OpenStreetMap-based map views, overlaid satellite imagery, crop-classification rasters with adjustable opacity, and drone-analysis tabs for UAV patrol paths and orthomosaic-derived products. We connect each application module to dedicated state and tool panels, so users can inspect the execution chain, intermediate outputs, and final geospatial products within the same workflow interface.

\begin{table*}[ht!]
  \caption{Agent performance on \textbf{L3-test-mini} full-season scenarios with A2A enabled.}
  \label{tab:a2a-on-results}
  \begin{center}
        \begin{tabular}{ll|cccccc|cccccc}
        \toprule        
        Agent & In-context  & Yield & Succ. & Path & KTC & Tkns/ & Tkns/ & Yield & Succ. & Path & KTC & Tkns/ & Tkns/ \\
        Routing & Learning  & Loss & Score & Corr. & Score & L3 sc. & Step & Loss & Score & Corr. & Score & L3 sc. & Step \\
        \midrule
        \rowcolor{black!10}
        & & \multicolumn{6}{c|}{\textit{Qwen3.6-35B-A3B}} & \multicolumn{6}{c}{\textit{DeepSeek-V4-Flash}}  \\
        \multirow{4}{*}{Direct}  & Zero Context
          & 27.3\% & 38.7\% & 22.9\% & 88.9\% & 3.38M & 31.9k
          & 16.3\% & 43.9\% & 26.9\% & 87.4\% & 5.03M & 39.4k \\
        & LLM-as-an-Expert
          & 9.8\% & 47.8\% & 25.0\% & 89.8\% & 4.12M & 35.2k
          & 8.5\% & 56.1\% & 25.2\% & 90.5\% & 5.40M & 39.8k \\
        & Skills Library
          & 1.6\% & 61.4\% & 30.5\% & 93.8\% & 4.01M & 38.1k
          & 1.9\% & 67.2\% & 32.8\% & 93.5\% & 7.29M & 50.9k \\
        & Expert Instruct
          & 4.0\% & 63.1\% & 29.5\% & 92.8\% & 2.94M & 29.0k
          & 1.8\% & 67.2\% & 30.0\% & 92.7\% & 4.50M & 36.5k \\
        \midrule
        \multirow{4}{*}{A2A} & Zero Context       & 43.3\% & 22.1\% & 15.3\% & 84.0\% & 1.26M & 12.1k & 54.4\% & 22.2\% & 14.6\% & 92.8\% & 4.97M & 25.3k  \\
        & LLM-as-an-Expert   & 24.6\% & 39.4\% & 21.2\% & 89.1\% & 1.61M & 10.7k & 33.6\% & 36.7\% & 16.7\% & 85.7\% & 2.16M & 11.9k \\
        & Skills Library     & 18.0\% & 51.7\% & 22.5\% & 90.8\% & 2.22M & 12.9k & 27.5\% & 46.5\% & 27.6\% & 94.3\% & 3.81M & 21.0k \\
        & Expert Instruct    & 31.8\% & 43.6\% & 21.4\% & 87.5\% & 1.10M & 10.9k & 9.5\%  & 42.4\% & 22.0\% & 87.5\% & 1.69M & 12.6k \\
        \bottomrule
        \end{tabular}
  \end{center}
\end{table*}

\begin{table*}[ht!]
  \caption{Agent performance on \textbf{L3-validation} full-season scenarios across library schemes. Flat structures correspond to directly retrieving against L3-level contexts, and hierarchical structures correspond to multi-level (L1, L2, L3) retrieval contexts.}
  \label{tab:qwen-level-results-app}
  \begin{center}
  \resizebox{1.0\textwidth}{!}{
        \begin{tabular}{lll|cccccc|cccccc}
        \toprule        
          Knowledge  & Library & Similarity & Yield & Succ. & Path & KTC & Tkns/ & Tkns/ & Yield & Succ. & Path & KTC & Tkns/ & Tkns/ \\
          Library & Structure & Mechanism & Loss & Score & Corr. & Score & L3 sc. & Call & Loss & Score & Corr. & Score & L3 sc. & Call \\
        \midrule
        \rowcolor{black!10}
        & & & \multicolumn{6}{c|}{\textit{Qwen3.6-35B-A3B}} & \multicolumn{6}{c}{\textit{DeepSeek-V4-Flash}}  \\
        Zero Context & -- & --
        & 21.6\% & 37.6\% & 22.2\% & 85.6\% & 3.54M & 33.0k
        & 22.3\% & 43.5\% & 27.8\% & 88.7\% & 5.12M & 38.2k \\
        Expert Instruct & -- & --
        & 3.1\% & 60.2\% & 26.7\% & 91.9\% & 3.55M & 31.1k
        & 1.3\% & 70.1\% & 28.6\% & 93.3\% & 5.74M & 40.3k \\
        \midrule
        \ub{GeoLLM-Engine}~\cite{singh2024geollm} 
          & \ub{Flat} & \ub{Text (Manual)}
          & \ub{3.0\%} & \ub{60.8\%} & \ub{34.3\%} & \ub{92.7\%} & \ub{4.74M} & \ub{40.4k} & \ub{3.7\%} & \ub{63.8\%} & \ub{34.4\%} & \ub{93.6\%} & \ub{6.19M} & \ub{44.6k} \\
        \ub{GeoFlow}~\cite{Bhattaram2025geoflow} 
          & \ub{Hierarchical} & \ub{Full-path (Manual)} 
          & \ub{2.9\%} & \ub{68.0\%} & \ub{34.8\%} & \ub{97.2\%} & \ub{4.57M} & \ub{39.5k} & \ub{2.8\%} & \ub{73.5\%} & \ub{41.1\%} & \ub{96.8\%} & \ub{7.60M} & \ub{47.4k} \\
          \midrule
        PowerChain~\cite{badmus2025powerchain__v2} 
          & Flat & Textual
          & 7.9\% & 59.1\% & 35.9\% & 94.6\% & 4.41M & 40.1k & 2.4\% & 71.9\% & 38.1\% & 94.5\% & 7.84M & 47.4k \\
        PowerDAG~\cite{badmus2026powerdagreliableagenticai} 
          & Flat & Full-path 
          & 7.8\% & 60.6\% & 34.7\% & 95.3\% & 4.44M & 40.2k & 4.6\% & 67.6\% & 38.5\% & 95.8\% & 7.45M & 44.2k \\
        HTAM~\cite{li2025designingdomainspecificagentshierarchical} 
          & Hierarchical & Full-path 
          & 3.5\% & 58.8\% & 31.0\% & 92.9\% & 4.73M & 40.4k & 8.1\% & 58.2\% & 32.1\% & 93.7\% & 5.65M & 44.9k \\
          \midrule
        FAIRY Skills Library 
          & Hierarchical & Text + Grouped Paths 
          & 3.5\% & 61.1\% & 32.0\% & 93.8\% & 5.11M & 41.6k & 3.0\% & 63.1\% & 32.4\% & 93.9\% & 6.04M & 46.1k \\
        \bottomrule
        \end{tabular}
        }
  \end{center}
\end{table*}

\begin{figure*}[t]
  \centering
  \includegraphics[width=\textwidth]{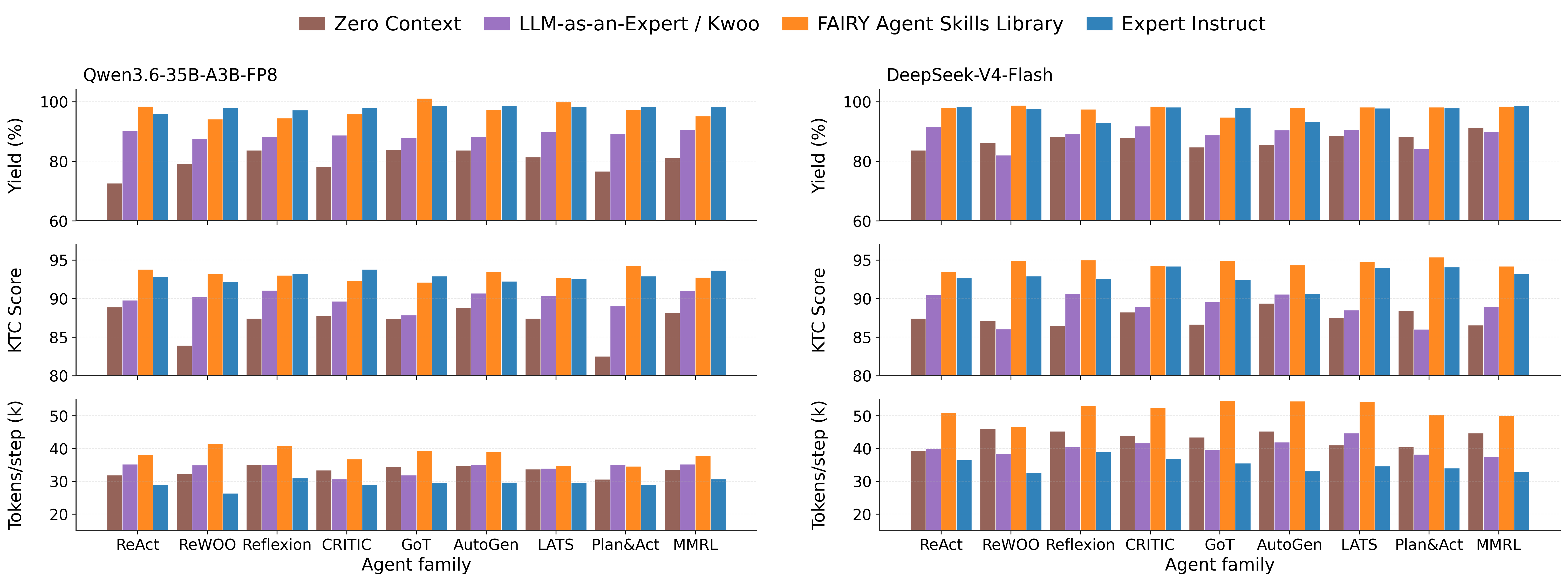}
  \caption{Agent-family and backbone comparison on \textbf{L3-mini} full-season scenarios. The left and right panels report Qwen3.6-35B-A3B-FP8 and DeepSeek-V4-Flash, respectively. For each agent family, colored bars denote different context settings, and rows show full-season yield, FAIRY KTC score, and token cost per step.}
  \label{fig:e3-agent-family-backbone}
  \Description{Figure summarizing agentic performance results across different controller and backbone implementations.}
\end{figure*}

\section{Results}

We organize results around the questions an operator would ask before
deploying: what makes agents reliable (context and skills), how reliability
scales with horizon, what multi-agent orchestration costs, which metric we
should trust, and what deployment costs.

\textbf{Expert context dominates the controller spread.} Table~\ref{tab:horizon-api-results} reports the held-out $N{=}70$ L3 evaluation under four in-context regimes for both backbones. Context is the dominant
lever. For Qwen, moving from Zero Context to Expert Instruct cuts Yield Loss
from $22.3\%$ to $4.6\%$ and raises KTC from $87.9\%$ to $92.1\%$; the Skills
Library regime is close behind ($4.9\%$ Yield Loss, $93.7\%$ KTC) without any
hand-written per-scenario policy. DeepSeek shows the same ordering (Zero
$\rightarrow$ Expert: $12.6\%\rightarrow2.9\%$ Yield Loss). Notably,
\emph{LLM-as-an-Expert} recovers only part of the gap (Qwen $15.3\%$ Yield Loss
versus $4.6\%$ for human instructions), indicating that agriculture-tuned LLM
context captures high-level decisions but not the full procedural discipline.

\begin{table*}[ht!]
  \caption{Local vLLM performance on atomic tasks (L1), episode chains (L2) and full-season scenarios (L3-mini) under expert-instructed context. Runtime (seconds) is measured on NVIDIA Thor.}
  \label{tab:qwen-vllm-horizon-results}
\begin{center}
        \begin{tabular}{llcccccccc}
        \toprule        
          Tasks & Model & Yield & Succ. & Path & KTC & Tkns/ & Tkns/ & Runtime/ & Runtime/ \\
          Level & Type & Loss & Score & Corr. & Score & Task & Step & Step  & Task  \\
        \midrule
        \multirow{6}{*}{L1} & Qwen3.6-35B-A3B-FP8      & 0.2\% & 85.5\% & 65.3\% & 98.9\% & 0.29M & 13.8k & 4.3s & 101.1s \\
        & Qwen3.6-27B-FP8 & 0.2\% & 87.7\% & 68.3\% & 99.0\% & 0.29M & 13.8k & 20.2s & 427.0s \\
        & Qwen3.5-9B & 1.7\% & 87.5\% & 67.1\% & 99.3\% & 0.39M & 16.6k & 12.8s & 312.6s \\
        & Qwen3.5-4B & 0.6\% & 85.7\% & 59.0\% & 98.3\% & 0.82M & 27.3k & 11.0s & 411.9s \\
        & Qwen3.5-2B & 17.6\% & 46.3\% & 13.1\% & 97.8\% & 7.04M & 56.6k & 2.3s & 735.7s \\
        & Gemma-4-31B-IT-NVFP4 & 0.1\% & 87.3\% & 69.1\% & 98.8\% & 0.31M & 15.3k & 16.7s & 339.5s \\
        \midrule
        \multirow{6}{*}{L2} & Qwen3.6-35B-A3B-FP8  & 1.1\% & 72.1\% & 49.2\% & 97.9\% & 0.87M & 25.2k & 4.4s & 164.7s \\
                   & Qwen3.6-27B-FP8      & 0.8\% & 70.1\% & 45.7\% & 97.6\% & 0.77M & 22.2k & 20.7s & 720.4s \\
                   & Qwen3.5-9B      & 1.2\% & 67.3\% & 47.2\% & 97.5\% & 0.85M & 26.6k & 15.7s & 510.6s \\
                   & Qwen3.5-4B      & 1.5\% & 67.5\% & 48.4\% & 96.4\% & 1.45M & 30.3k & 6.6s & 931.4s \\
                   & Qwen3.5-2B      & 11.7\% & 31.8\% & 11.2\% & 90.6\% & 5.53M & 48.4k & 2.5s & 518.4s \\
                   & Gemma-4-31B-IT-NVFP4      & 1.1\% & 69.6\% & 56.0\% & 98.6\% & 0.68M & 23.3k & 19.1s & 567.5s \\
        \midrule
        \multirow{11}{*}{L3} & Qwen3.6-35B-A3B-FP8  & 2.1\% & 66.6\% & 28.1\% & 94.2\% & 3.77M & 31.9k & 3.3s & 407.1s \\
                             & Qwen3.6-27B-FP8      & 2.3\% & 66.6\% & 35.1\% & 93.1\% & 2.56M & 26.1k & 14.7s & 1451.8s \\
                             & Qwen3.5-9B      & 3.0\% & 54.6\% & 26.2\% & 93.5\% & 2.60M & 29.5k & 11.1s & 981.9s \\
                             & Qwen3.5-4B      & 5.0\% & 50.7\% & 22.2\% & 88.1\% & 4.07M & 36.5k & 7.6s & 861.2s \\
                             & Gemma-4-31B-IT-NVFP4      & 2.1\% & 53.9\% & 26.6\% & 92.0\% & 1.93M & 26.3k & 17.1s & 1267.2s \\
        & Gemma-4-26B-A4B-NVFP4  & 2.4\% & 45.6\% & 18.9\% & 88.2\% & 1.53M & 25.1k & 3.5s & 222.5s \\
        & Gemma-4-26B-A4B-NVFP4-assistant   & 12.0\% & 42.6\% & 16.2\% & 86.1\% & 1.79M & 31.2k & 7.5s & 444.1s \\
        & Gemma-4-12B-it   & 0.9\% & 59.5\% & 29.1\% & 92.4\% & 2.58M & 30.6k & 12.3s & 1039.9s \\
        & Gemma-4-12B-it-assistant   & 7.1\% & 58.3\% & 27.8\% & 91.7\% & 3.20M & 35.0k & 9.0s & 823.5s \\
        & Gemma-4-E4B-it   & 31.2\% & 41.7\% & 19.5\% & 80.8\% & 1.29M & 23.6k & 6.4s & 382.8s \\
        & Gemma-4-E2B-it  & 40.0\% & 31.3\% & 9.8\% & 67.5\% & 1.76M & 26.2k & 3.4s & 238.9s \\
        \bottomrule
        \end{tabular}
  \end{center}
\end{table*}

\begin{figure*}[ht!]
  \centering
  \includegraphics[width=\linewidth]{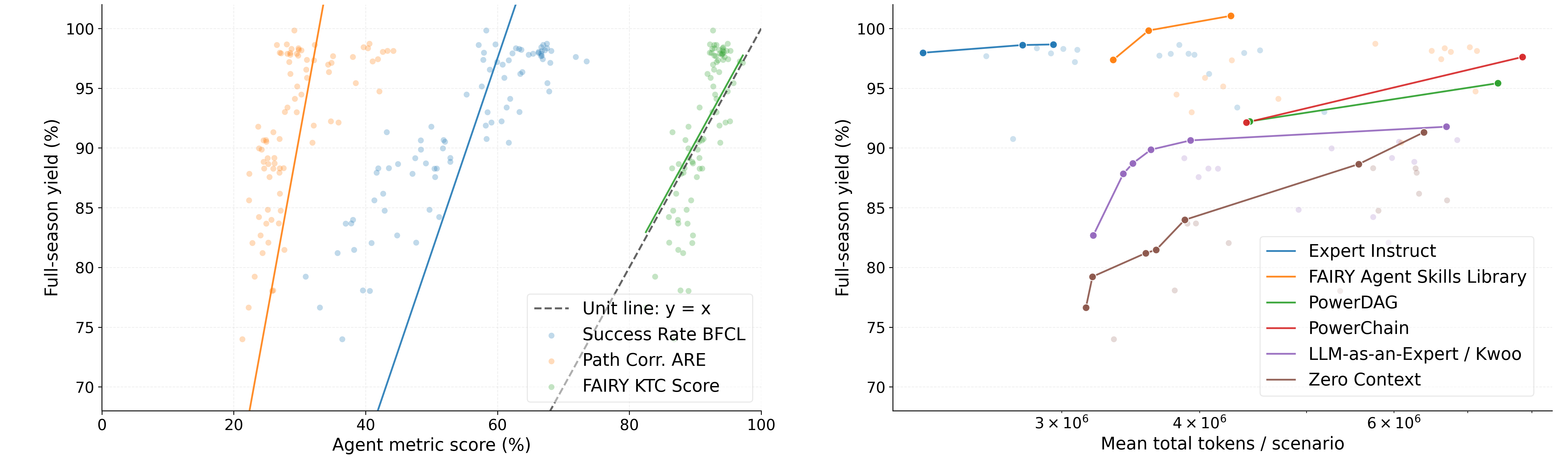}
  \caption{Left: Assessing full-path agent metrics vs. biological yield preservation. Right: Evaluating the cost vs. downstream objective performance (yield) tradeoff across scenario runs.}
  \Description{Figure summarizing the alignment of agentic evaluation metrics with  downstream agronomic performance.}
\label{fig:metrics}
\end{figure*}

\textbf{Knowledge-library structure and retrieval.}
Table~\ref{tab:qwen-level-results-app} ablates how the knowledge library is
\emph{organized} (flat vs.\ tier-grouped) and how entries are \emph{ranked}
(manual, text similarity, path similarity) on the L3-mini set. Tier-grouped
organization with similarity-based retrieval gives the strongest correctness
(KTC up to $97.2\%$ for Qwen, $96.8\%$ for DeepSeek) and the best task success,
while flat/manual retrieval is weaker and more variable. The effect on yield is
smaller than the context effect, which is
expected: retrieval quality refines \emph{how} an already-grounded agent acts,
whereas context grounding determines whether it acts correctly at all.

\textbf{Horizon analysis: short tasks are nearly solved, full seasons are not.}
Table~\ref{tab:horizon-api-results} reports atomic (L1) and episodic (L2) tasks.
Under expert context, both horizons are essentially solved: L1 reaches
$\geq$99\% KTC with near-zero yield loss, and L2 reaches $\geq$98\% KTC with
$<$1\% yield loss. The contrast with the L3 results is the central horizon finding: short tasks
mostly test whether the scaffold can select tools and satisfy local
preconditions, whereas full-season scenarios test whether decisions remain
coherent after their effects propagate through soil state, growth, stress
accumulation, treatment residuals, harvest timing, and storage. Yield loss is
where the horizon bites: it is small on L1/L2 but reaches double digits on L3
under zero context.

\begin{figure*}[ht!]
  \centering
  \includegraphics[width=\linewidth]{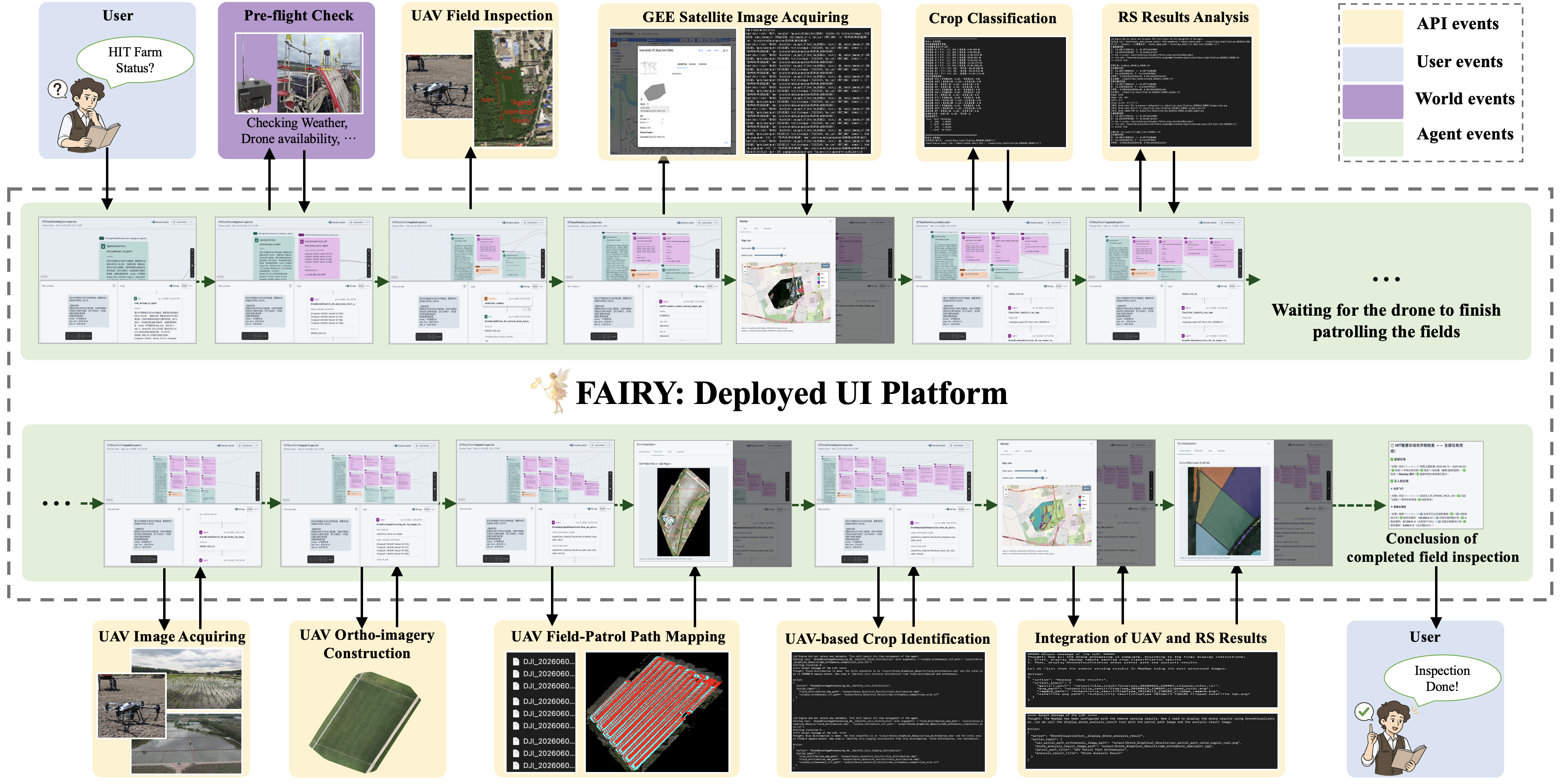}
  \caption{FAIRY Operational Demonstration:
  drone patrol and Sentinel-2 crop classification coordinated through the
  event-driven runtime, rendered as a geospatial inspection result in the system
  UI.}
  \label{fig:FAIRY_demo}
  \Description{Operational demonstration of the proposed FAIRY framework on our research farm.}
\end{figure*}

\textbf{Multi-agent orchestration introduces coordination cost.}
Table~\ref{tab:a2a-on-results} compares direct tool access against A2A routing,
where the controller coordinates with weather, sensing, machinery, and
operations specialists. Averaged over contexts, A2A degrades both correctness
and yield on both backbones: Yield Loss rises from $7.1\%$ to $31.2\%$
(DeepSeek) and from $10.7\%$ to $29.4\%$ (Qwen).
Token cost per scenario drops because work is offloaded to specialists, but the
main controller frequently loses track of what each specialist observed or
executed. The effect is controller-dependent (tree search is comparatively
robust), consistent with coordination-overhead observations in other
expert-level multi-agent studies~\cite{lee2025multiagent, badmus2025powerchain__v2}. Early analysis suggests errors due to task decomposition issues, so we plan a detailed failure taxonomy as future work.

\textbf{Which trace metric predicts yield?} Figure~\ref{fig:metrics} (left) and Table~\ref{tab:metric_yield_alignment} regress each metric against biological yield preservation across the full sweep. Overall, we observe that KTC is the best-calibrated predictor as it lies almost on
the unit line. Moreover, we note that BFCL's high $R^2$ performance comes from robustly identifying catastrophic ``never harvested'' runs (zero recovered yield), rather than grading quality among completing
runs: it detects failure but may be poorly calibrated. Overall, the results show that existing practices of order-only and exact-match metrics might not fully capture downstream tasks. Overall, we consider the following for our practical deployment and future work: report \textbf{KTC} for trace-level
correctness and \textbf{Yield Loss} for outcome, and treat exact-match success
as a failure detector rather than a quality measure.

\begin{table}[t!]
\centering
\caption{Alignment between agent-evaluation metrics and the downstream agronomic objective, measured by yield. Lower RMSE and slope closer to 1 indicate better calibration, while higher $R^2$ indicates stronger explanatory fit.}
\label{tab:metric_yield_alignment}
\small
\begin{tabular}{lccc}
\toprule
\textbf{Full-path agent metric} & \textbf{RMSE} & \textbf{Slope $a$} & \textbf{$R^2$} \\
\midrule
Success rate BFCL~\cite{patil2025bfcl} & 37.1 & 1.63 & 0.81 \\
Path correctness ARE~\cite{froger2025are} & 62.5 & 3.04 & 0.41 \\
FAIRY KTC score & 4.2 & 1.01 & 0.70 \\
\bottomrule
\end{tabular}
\end{table}

\textbf{Edge deployment profiling.}
Because field deployment cannot assume frontier-API availability, we assess performance and cost under edge deployment considerations. Table~\ref{tab:qwen-vllm-horizon-results} shows that local vLLM execution can support L1 and L2 farm tasks with low yield loss across several mid-size models, but full-season L3 scenarios separate models more clearly: Qwen3.6-35B-A3B-FP8 and Qwen3.6-27B-FP8 keep yield loss near 2\%, while smaller or assistant-tuned variants degrade sharply. In practice, this suggests that edge deployment is feasible for farm-agent evaluation, but full-season autonomy still requires models with enough planning capacity.

\textbf{Per-controller robustness.}
Figure~\ref{fig:e3-agent-family-backbone} summarizes the performance across the 9 different agentic back-ends on the L3-mini set across the four in-context regimes. Overall, we observe that in-context operational grounding is the dominant factor across backbones: moving from zero context to the skills library or expert instruction reduces average yield loss from 19.9\% to 2.9\%/2.1\% for Qwen3.6-35B-A3B and from 12.8\% to 2.2\%/3.0\% for DeepSeek-V4-Flash. In practice, this means that the farm deployment \textbf{cannot} rely on generic, state-of-the-art agent scaffolds alone; agents need explicit agronomic skills or expert operational context to preserve yield under full-season decision sequences.

\section{FAIRY Operational Demonstration}

We demonstrate FAIRY on a real-farm inspection episode at the HIT
smart-agriculture site. As shown in Figure~\ref{fig:FAIRY_demo}, the demonstration focuses on the following L2 episode: a user issues a field-monitoring request, and the agent must coordinate satellite and drone observations, crop-identification tools, post-processing, and the FAIRY UI to return an interpretable inspection result.  The integrated episode exercises the full observation workflow: the agent first
triggers the drone API, retrieves and classifies satellite imagery while the drone is in flight; it then retrieves the drone result, generates the orthomosaic, plots the patrol path, runs field/crop
analysis, integrates satellite and drone findings, and updates the UI display
tool to render base-map context, classification results, the patrol path, and
image-analysis products. Overall, this demonstration provides us with 
a working deployment prototype and end-to-end path from user request, through event-driven agent execution, to farm-facing geospatial visualization.

\section{Discussion}

\textbf{From expert instructions to reusable skills.} Encoding tacit expert
knowledge as scenario-specific instructions is effective for evaluation but
does not scale: each new task or seasonal edge case would need another written
policy. Organizing this knowledge into retrievable, composable
\textbf{skill libraries} is the scalable alternative, consistent with the use
of structured knowledge pools in power-grid~\cite{badmus2025powerchain__v2} and
Earth-observation~\cite{Bhattaram2025geoflow,
shabbir2026thinkgeoevaluatingtoolaugmentedagents,
chen2026canglingknowflowunifiedknowledgeandflowfusedagent,
feng2026earthagentunlockinglandscapeearth__v2} agents; our library ablation is a first step in this direction.

\textbf{Full-path evaluation paired with farm objectives.} A planting,
irrigation, or spraying decision can look locally plausible while still causing
downstream yield loss, so the downstream objective must be evaluated alongside
the action trace. KTC and Yield Loss diagnose different failure modes: a
high-KTC, high-yield-loss run shows that small operational differences can have
physical consequences, while a lower-KTC, low-yield-loss run shows deviation
from the reference trace that nonetheless preserved the outcome. Our
metric-calibration study sharpens this: among
trace metrics, temporally grounded correctness tracks yield, whereas flat
order metrics and success measures do not. We note that the current analysis aggregates results at the scenario level. A natural next step is ridge-level spatial analysis, including whether errors cluster around sparsely instrumented ridges, low-observability zones, or operations that depend on UAV and satellite coverage.

\textbf{Agent orchestration should preserve operational state.} Planning,
memory, retrieval, verification, and specialist decomposition can improve the
execution trace, but they do not remove the need to maintain agronomic and
operational assumptions across time. Splitting the farm interface into
specialists matches the structure of real farm systems, yet introduces
state-sharing requirements; if the controller loses track of what each
specialist observed or executed, orchestration becomes error-prone. The design requirement is not only better decomposition but preserving shared farm state, timing constraints, and task context across agent boundaries.

\textbf{LLM-as-an-Expert context.} Even when a domain
LLM~\cite{wang2025agrikwoo} is given the scenario, tools, world state, and a
matched response template, it does not fully match human experts on procedural
ordering. This does not make the domain LLM unhelpful, as it recovers
high-level crop-management choices, but it motivates structured knowledge
libraries that complement both LLM-distilled and human-written guidance.

\textbf{Quality of human-annotated solutions.} Because every trace
metric compares against a human-oracle workflow, the oracle must be a fixed,
shared artifact; regenerating it across software versions or machines
introduces drift that silently changes every metric. We recommend version-controlling and serializing the oracle workflows alongside each scenario. Moreover, a timing-aware metric can only be validated where mistiming actually costs yield, making such information available in the oracle traces particularly important. The next stage of deployment is to run an agent-managed plot alongside the human-operated workflow and compare both operational traces and harvested outcomes under real seasonal conditions.

\section{Conclusion}

We presented FAIRY, a deployable smart-agriculture agentic engine, and used it
for a full-season spatiotemporal evaluation of contemporary agent practices on
an operating soybean research farm. Expert context and retrievable agronomic
skills are the dominant levers for long-horizon reliability; multi-agent
orchestration and deployment cost are practical constraints; and among
trace-level metrics, temporally grounded correctness is the one that tracks
real yield. These results inform our next stage: operating a dedicated plot
under agent management alongside the human-operated farm. We hope FAIRY provides
the community a deployable, spatiotemporally grounded reference point for
evaluating agents in other physical-process settings. Our entire working prototype can be found here: \url{https://github.com/Fengrui-Lab/FAIRY}

\begin{acks}
This work was supported by the National Natural Science Foundation of China (grant No. 62350710797). We gratefully acknowledge the support of the National Key Research and Development Program [2025YFE0209200] and the Key Research and Development Program of Heilongjiang Province, China [2024ZX01A07, JD2023GJ01]. This work was also supported by the NSFC grant (No. 42471362). DS gratefully acknowledges the support of the NSFC Excellent Young Scientists Fund Program (Overseas).
\end{acks}

\bibliographystyle{ACM-Reference-Format}
\bibliography{main}

\end{document}